\PassOptionsToPackage{table}{xcolor}

\documentclass{article} 
\usepackage{nips15submit_e,times}

\usepackage{graphicx}
\usepackage{arydshln}
\usepackage{subfig}
\usepackage{epsfig}
\usepackage{amsmath}
\usepackage{amssymb}
\usepackage{textcomp}
\usepackage{sidecap}
\sidecaptionvpos{figure}{c}
\usepackage{color}
\usepackage{wrapfig}
\usepackage[table]{xcolor}

\nipsfinalcopy

\sidecaptionvpos{figure}{b}

\usepackage[pagebackref=false,breaklinks=true,letterpaper=true,colorlinks,bookmarks=false]{hyperref}

\begin{document}

\title{Estimating visual classifiers without data}
\title{Learning visual biases from human imagination}

\author{Carl Vondrick \hspace{1em} Hamed Pirsiavash$\dagger$ \hspace{1em} Aude Oliva \hspace{1em} Antonio Torralba\\
Massachusetts Institute of Technology \hspace{1em} $\dagger$University of Maryland, Baltimore County\\
\texttt{\{vondrick,oliva,torralba\}@mit.edu} \hspace{1em} \texttt{hpirsiav@umbc.edu}
}

%
%

\maketitle

\begin{abstract}

Although the human visual system can recognize many concepts under challenging
conditions, it still has some biases. In this paper, we investigate whether we
can extract these biases and transfer them into a machine recognition system.
We introduce a novel method that, inspired by well-known tools in human
psychophysics, estimates the biases that the human visual system might use for
recognition, but in computer vision feature spaces.   Our experiments are
surprising, and suggest that classifiers from the human visual system can be
transferred into a machine with some success. Since these classifiers seem to
capture favorable biases in the human visual system, we further present an SVM
formulation that constrains the orientation of the SVM hyperplane to agree with
the bias from human visual system. Our results suggest that transferring this
human bias into machines may help object recognition systems generalize across
datasets and perform better when very little training data is available.

\end{abstract}

\section{Introduction}
\label{sec:intro}

%
%


Computer vision researchers often go through great lengths to remove dataset biases
from their models \cite{torralba2011unbiased,kulis2011you}. However, not all biases are adversarial. Even natural recognition systems,
such as the human visual system, have biases. Some of the most well known
human biases, for example, are the canonical perspective (prefer to see objects from
a certain perspective) \cite{palmer1981canonical} and Gestalt laws of grouping
(tendency to see objects in collections of parts) \cite{ellis1999source}.

We hypothesize that biases in the human visual system can be beneficial for
visual understanding. Since recognition is an underconstrained problem, the
biases that the human visual system developed may provide useful priors for
perception. In this paper, we develop a novel method to learn some biases from
the human visual system and incorporate them into computer vision systems.

\begin{figure*}[t]
    \centering
    \includegraphics[width=\linewidth]{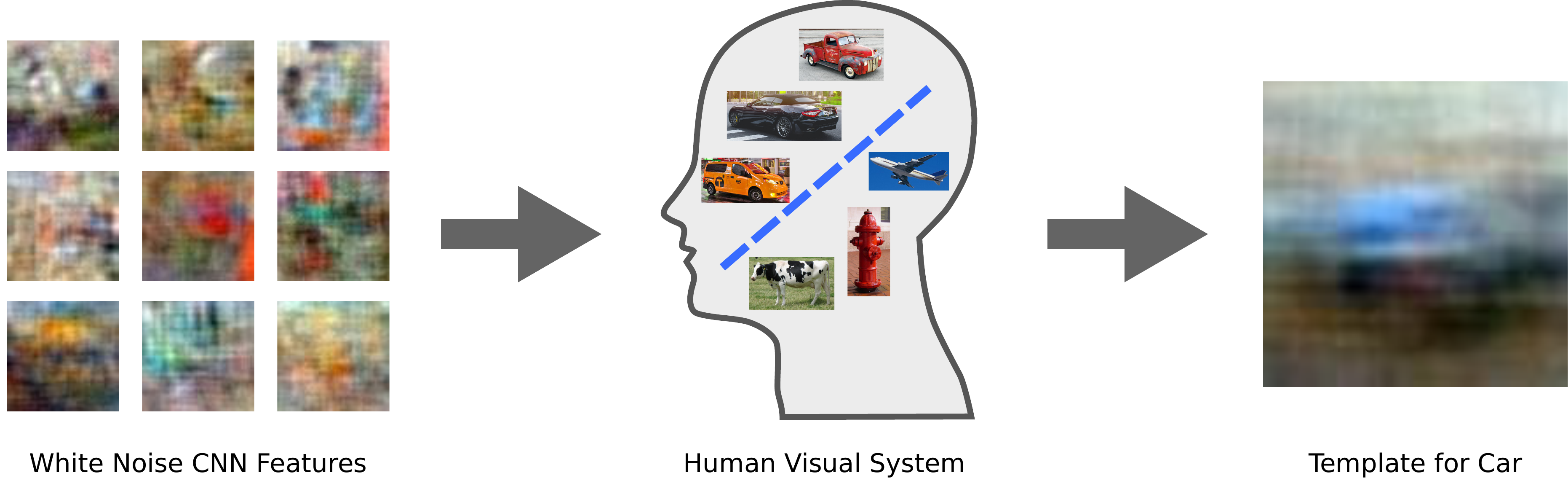}
\caption{Although all image patches on the left are just noise, when
we show thousands of them to online workers and ask them to find ones that look
like cars, a car emerges in the average, shown on the right.  This
noise-driven method is based on well known tools in human psychophysics that
estimates the biases that the human visual system uses for
recognition. We explore how to transfer these biases into a machine.}

\label{fig:teaser}
\vspace{-2em}
\end{figure*}

We focus our approach on learning the biases that people may have for the
appearance of objects.  To illustrate our method, consider what may seem like
an odd experiment. Suppose we sample i.i.d.\ white noise from a standard normal
distribution, and treat it as a point in a visual feature space, e.g.\ CNN or
HOG.  What is the chance that this sample corresponds to visual features of a
car image? Fig.\ref{fig:teaser}a visualizes some samples
\cite{vondrick2013hoggles} and, as expected, we see noise.  But, let us not
stop there. We next generate one hundred fifty thousand points from the same
distribution, and ask workers on Amazon Mechanical Turk to classify
visualizations of each sample as a car or not. Fig.\ref{fig:teaser}c visualizes
the average of visual features that workers believed were cars. Although our
dataset consists of only white noise, a car emerges!

Sampling noise may seem unusual to computer vision researchers, but a similar
procedure, named classification images, has gained popularity in human
psychophysics \cite{ahumada1996perceptual} for estimating an approximate
template the human visual system internally uses for recognition
\cite{lovell1971stimulus,beard1998technique}. In the procedure, an observer
looks at an image perturbed with random noise and indicates whether they
perceive a target category.  After a large number of trials, psychophysics
researchers can apply basic statistics to extract an approximation of the
internal template the observer used for recognition. Since the procedure is
done with noise, the estimated template reveals some of the cues that the human
visual system used for discrimination.

We propose to extend classification images to estimate biases from the human
visual system. However, our approach makes two modifications. Firstly, we
estimate the template in state-of-the-art computer vision feature spaces
\cite{dalal2005histograms,krizhevsky2012imagenet}, which allows us to
incorporate these biases into learning algorithms in computer vision systems.
To do this, we take advantage of algorithms that invert visual features back to
images \cite{vondrick2013hoggles}. By estimating these biases in a feature space,
we can learn biases for how humans may correspond mid-level features, such as shapes and colors, with objects.
To our knowledge, we are the first to
estimate classification images in vision feature spaces.  Secondly, we want our
template to be biased by the human visual system and not our choice of dataset.
Unlike classification images, we do not perturb real images; instead
our approach only uses visualizations of feature space noise to estimate the
templates. We capitalize on the ability of people to discern visual objects
from random noise in a systematic manner \cite{gosselin2003superstitious}.

\section{Related Work}


\textbf{Mental Images}: Our methods build upon work to extract mental images
from a user's head for both general objects \cite{ferecatu2009statistical},
faces \cite{mangini2004making}, and scenes \cite{greene2014visual}. However,
our work differs because we estimate mental images in state-of-the-art computer
vision feature spaces, which allows us to integrate the mental images into a
machine recognition system.

\textbf{Visual Biases:} Our paper studies biases in the human visual system similar to \cite{palmer1981canonical,ellis1999source},
but we wish to transfer these biases into a computer recognition system. We extend ideas \cite{mezuman2012learning} to use computer vision
to analyze these biases. Our work is also closely related to dataset biases \cite{torralba2011unbiased,ponce2006dataset}, which motivates
us to try to transfer favorable biases into recognition systems.

\textbf{Human-in-the-Loop:} The idea to transfer biases from the human
mind into object recognition is inspired by many recent works that puts a human
in the computer vision loop \cite{branson2010visual,parikh2011human},
trains recognition systems with active learning
\cite{vijayanarasimhan2011large}, and studies crowdsourcing
\cite{von2006peekaboom,sorokin2008utility}.  The
primary difference of these approaches and our work is, rather than using
crowds as a workforce, we want to extract biases from the worker's visual systems.

\textbf{Feature Visualization:} Our work explores a novel application of feature visualizations \cite{weinzaepfel2011reconstructing,vondrick2013hoggles,mahendran2014understanding}. Rather than
using feature visualizations to diagnose computer vision systems, we use them to inspect and learn biases in the human visual system.

\textbf{Transfer Learning:} We also build upon methods in transfer learning to
incorporate priors into learning algorithms. A common transfer learning
method for SVMs is to change the regularization term $||w||_2^2$ to
$||w-c||_2^2$ where $c$ is the prior \cite{salakhutdinov2011learning,yang2007adapting}. However,
this imposes a prior on both the norm and orientation of $w$. In our case, since the
visual bias does not provide an additional prior on the norm, we
present a SVM formulation that constrains only the orientation of $w$
to be close to $c$. Our approach extends sign constraints on
SVMs \cite{epshteyn2005rotational}, but instead enforces orientation constraints.
Our method enforces a hard orientation constraint, which builds on soft orientation constraints \cite{aytar2011tabula}.



%

\section{Classification Images Review}

The procedure \emph{classification images} is a popular method in human
psychophysics that attempts to estimate the internal template that the human
visual system might use for recognition of a category
\cite{lovell1971stimulus,beard1998technique}.  We review
classification images in this section as it is the inspiration for our method.

The goal is to approximate the template $\tilde{c} \in \mathbb{R}^d$ that a human
observer uses to discriminate between two classes $A$ and $B$, e.g.\ male vs.\
female faces, or chair vs.\ not chair.  Suppose we have intensity images $a \in A
\subseteq \mathbb{R}^d$ and $b \in B \subseteq \mathbb{R}^d$. If we sample
white noise $\epsilon \sim \mathcal{N}(0^d, I_d)$ and ask an observer to
indicate the class label for $a+\epsilon$, most of the time the observer will
answer with the correct class label $A$. However, there is a chance that
$\epsilon$ might manipulate $a$ to cause the observer to mistakenly label
$a+\epsilon$ as class $B$. 

The insight into classification images is that, if we perform a large number of
trials, then we can estimate a decision function $f(\cdot)$ that discriminates
between $A$ and $B$, but makes the same mistakes as the observer.  Since
$f(\cdot)$ makes the same errors, it provides an estimate of the 
template that the observer internally used to discriminate $A$ from $B$. By analyzing this model, we can
then gain insight into how a visual system might recognize different categories. 

Since psychophysics researchers are interested in models that are
interpretable, classification images are often linear approximations
of the form $f(x; \tilde{c}) = \tilde{c}^T x$. The template $\tilde{c} \in \mathbb{R}^d$ can be estimated
in many ways, but the most common is a 
sum of the stimulus images:
\begin{align}
\tilde{c} &= \left(\mu_{AA} + \mu_{BA}\right) - \left(\mu_{AB} + \mu_{BB}\right)
\label{eqn:ci}
\end{align}
where $\mu_{XY}$ is the average image where the true class is $X$ and the observer predicted class $Y$. The template $c$ is fairly intuitive: it will have large positive value on locations that the observer used to predict $A$, and large negative value for locations correlated with predicting $B$.
Although classification images is simple, this procedure has led
to insights in human perception.  For example,
\cite{sekuler2004inversion} used classification images to study face processing strategies in the
human visual system.  For a complete analysis of
classification images, we refer readers to review articles
\cite{murray2011classification,eckstein2002classification}.

%
%
%


\section{Estimating Human Biases in Feature Spaces}

\sidecaptionvpos{figure}{c}
\begin{SCfigure}[1][t]
\centering 
\subfloat[RGB]{
\includegraphics[width=0.15\linewidth]{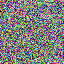}
}
\subfloat[HOG]{
\includegraphics[width=0.15\linewidth]{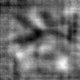}
}
\subfloat[CNN]{
\includegraphics[width=0.15\linewidth]{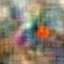}
}
\caption{We visualize white noise in RGB and feature spaces. To visualize white noise features,
we use feature inversion algorithms \cite{vondrick2013hoggles}. White noise in feature space has correlations in image space that
white noise in RGB does not. We capitalize on this structure to estimate visual biases 
in feature space without using real images.}
\label{fig:whitenoise}
\end{SCfigure}
\sidecaptionvpos{figure}{b}

Standard classification images is performed with perturbing real images with
white noise.  However, this approach may negatively bias the template by the
choice of dataset.  Instead, we are interested in estimating templates that
capture biases in the human visual system and not datasets.

We propose to estimate these templates by only sampling white noise (with no
real images). Unfortunately, sampling just white noise in RGB is extremely
unlikely to result in a natural image (see Fig.\ref{fig:whitenoise}a). To
overcome this, we can estimate the templates in feature spaces \cite{dalal2005histograms,krizhevsky2012imagenet} used in
computer vision.  Feature spaces encode higher abstractions of images (such as
gradients, shapes, or colors). While sampling white noise in feature space may
still not lay on the manifold of natural images, it is more likely to capture
statistics relevant for recognition.
Since humans cannot directly interpret abstract feature spaces, we can use
feature inversion algorithms \cite{vondrick2013hoggles,weinzaepfel2011reconstructing}
to visualize them.




Using these ideas, we first sample noise from a
zero-mean, unit-covariance Gaussian distribution $x \sim \mathcal{N}(0_d,
I_d)$. We then invert the noise feature $x$ back to an image $\phi^{-1}(x)$
where $\phi^{-1}(\cdot)$ is the feature inverse. By instructing people to
indicate whether a visualization of noise is a target category or not, we can
build a linear template $c \in \mathbb{R}^d$ that approximates people's
internal templates:
\begin{align}
c &= \mu_{A} - \mu_{B}
\label{eqn:cihog}
\end{align}
where $\mu_A \in \mathbb{R}^d$ is the average, in feature space, of white noise
that workers incorrectly believe is the target object, and similarly $\mu_B
\in \mathbb{R}^d$ is the average of noise that workers believe is
noise.

Eqn.\ref{eqn:cihog} is a special case of the original classification images
Eqn.\ref{eqn:ci} where the background class $B$ is white noise and the positive
class $A$ is empty. Instead, we rely on humans to hallucinate objects in noise
to form $\mu_A$.
Since we build these biases with only white Gaussian noise and no real images, 
our approach may be robust to many issues in dataset bias 
\cite{torralba2011unbiased}.  Instead, templates from our method
can inherit the biases for the appearances of objects present in the human visual system, which we suspect
provides advantageous signals about the visual world. 

In order to estimate $c$ from noise, we need to perform many trials, which we can conduct
effectively on Amazon Mechanical Turk \cite{sorokin2008utility}. We
sampled $150,000$ points from a standard normal
multivariate distribution, and inverted each sample with the feature inversion algorithm from HOGgles
\cite{vondrick2013hoggles}. We then instructed workers to indicate whether
they see the target category or not in the visualization.  Since we found that
the interpretation of noise visualizations depends on the scale, we show the
worker three different scales.  We paid workers $10$\textcent\ to label $100$
images, and workers often collectively solved the entire batch
in a few hours. In order to assure quality, we occasionally
gave workers an easy example to which we knew the answer, and only retained
work from workers who performed well above chance. We only used the easy
examples to qualify workers, and discarded them when computing the final
template.

\begin{figure*}[t]
\centering
\includegraphics[trim=0em 0em 0em 0em,clip,width=\linewidth]{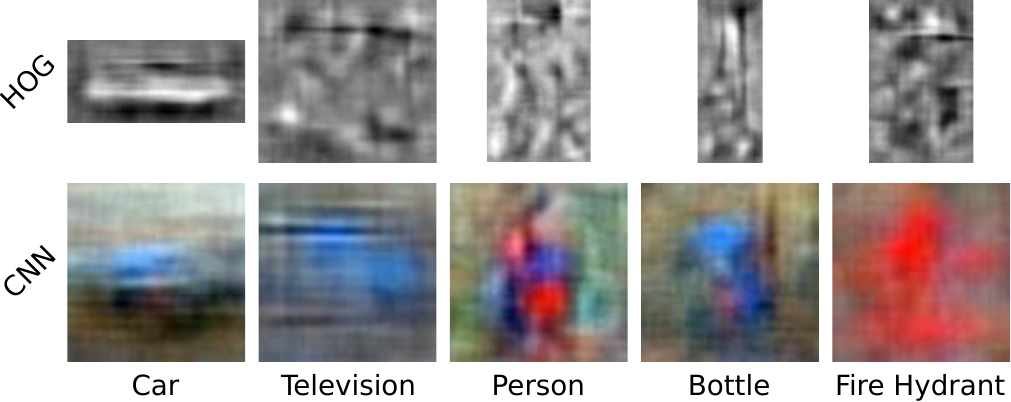}

\caption{We visualize some biases estimated from trials by Mechanical Turk
workers. }

\vspace{-1em}

\label{fig:examples}
\end{figure*}



\section{Visualizing Biases} 

Although subjects are classifying zero-mean, identity covariance white Gaussian
noise with no real images, objects can emerge after many trials. To show this,
we performed experiments with both 
HOG
\cite{dalal2005histograms} and the last convolutional layer (\texttt{pool5}) of a
convolutional neural network (CNN) trained on ImageNet
\cite{krizhevsky2012imagenet,deng2009imagenet}
for several common object
categories.  We visualize some of the templates from our method in
Fig.\ref{fig:examples}. Although the templates are blurred, they seem to show
significant detail about the object.  For example, in the car template, we can
clearly see a vehicle-like object in the center sitting on top of a dark road
and lighter sky.  The television template resembles a rectangular structure,
and the fire hydrant templates reveals a red hydrant with two arms on the side.
The templates seem to contain the canonical perspective of objects
\cite{palmer1981canonical}, but also extends them with color and shape biases.



\begin{SCfigure}[1.5][tb]
\centering
\subfloat[India]{
\includegraphics[width=0.2\linewidth]{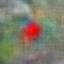}
}
\hspace{0.5em}
\subfloat[United States]{
\includegraphics[width=0.2\linewidth]{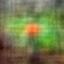}
}
\caption{We grouped users by their geographic location (US or India) and instructed each group to classify CNN noise as a sports ball or not,
which allows us to see how biases can vary by culture. Indians seem to imagine
a red ball, which is the standard color for a cricket ball and the predominant sport in India. Americans seem to imagine
a brown or orange ball, which could be an American football or basketball, both popular sports in the U.S.}
\label{fig:mental-printing-2}
\end{SCfigure}

In these visualizations, we have assumed that all workers on Mechanical Turk share
the same appearance bias of objects. However, this assumption is not necessarily true. 
To examine this, we instructed workers on Mechanical
Turk to find ``sport balls'' in CNN noise, and clustered workers by their
geographic location. Fig.\ref{fig:mental-printing-2} shows the templates 
for both India and the United States. Even though both sets of workers
were labeling noise from the same distribution, Indian workers seemed to imagine red balls, while
American workers tended to imagine orange/brown balls. Remarkably, the most
popular sport in India is cricket, which is played with a red ball, and popular
sports in the United States are American football and basketball, which are
played with brown/orange balls.
We conjecture that Americans and Indians may have different mental images of
sports balls in their head and the color is influenced by popular sports in
their country.  This effect is likely attributed to phenomena in social
psychology where human perception can be influenced by culture
\cite{chua2005cultural,blais2008culture}. Since environment plays a role in the
development of the human vision system, people from different cultures likely
develop slightly different images inside their head.

\section{Leveraging Humans Biases for Recognition}
\label{sec:experiments}

If the biases we learn are beneficial for recognition, then we would expect
them to perform above chance at recognizing objects in real images. To evaluate
this, we use the visual biases $c$ directly as a classifier for object
recognition. We quantify their performance on object classification in
real-world images using the PASCAL VOC 2011 dataset \cite{Everingham10},
evaluating against the validation set. Since PASCAL VOC does not have a fire
hydrant category, we downloaded $63$ images from Flickr with fire hydrants and
added them to the validation set. We report performance as the average precision on a
precision-recall curve.



%

\setlength{\tabcolsep}{3pt}

\sidecaptionvpos{figure}{c}
\begin{figure}[b]
\centering
\includegraphics[width=0.5\linewidth]{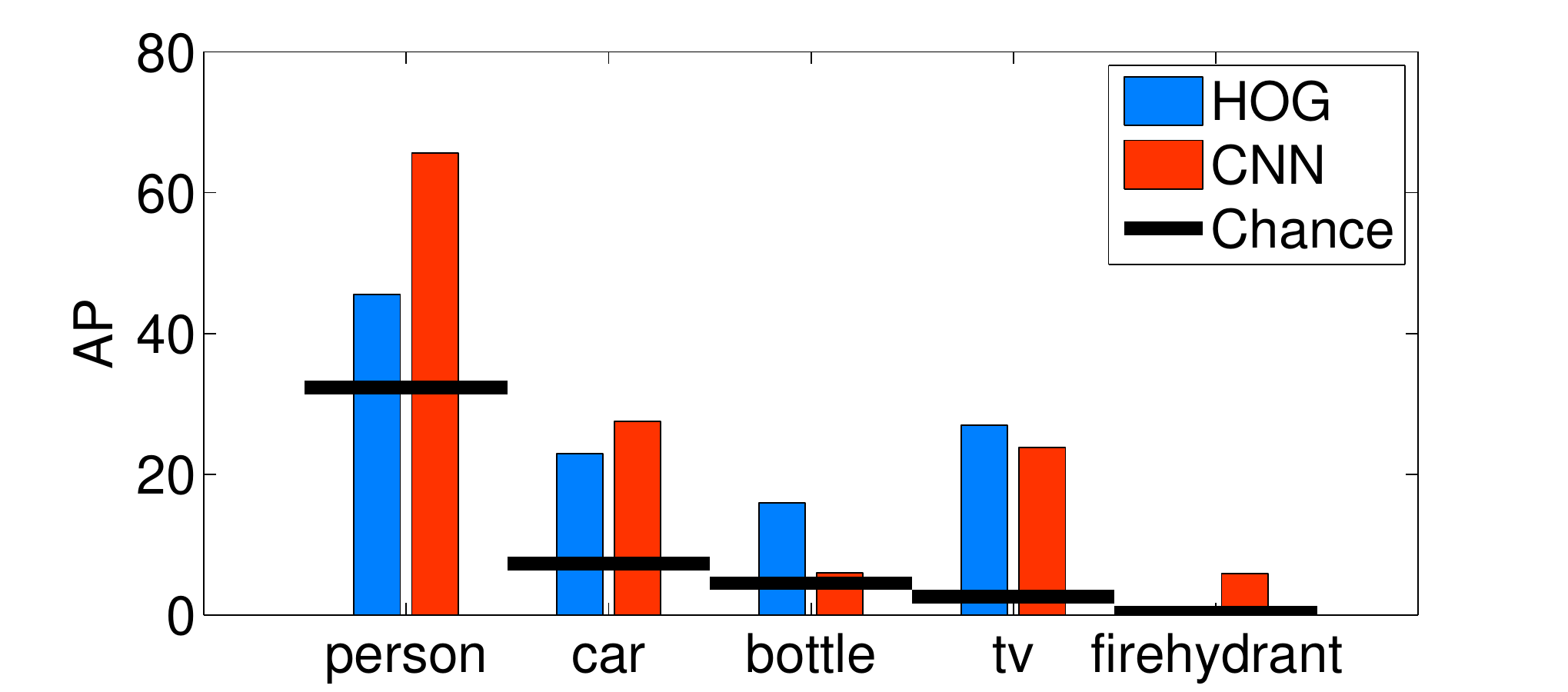}
\begin{tabular}[b]{r | c c c c c}
& car & person & f-hydrant & bottle & tv \\
\hline
HOG & 22.9 & 45.5 & 0.8 & 15.9 & 27.0 \\
CNN & 27.5 & 65.6 &  5.9 & 6.0  &23.8 \\
\hline
Chance & 7.3 & 32.3 & 0.3 & 4.5 & 2.6  \\
\multicolumn{3}{c}{} \\
\multicolumn{3}{c}{} \\
\end{tabular}
\vspace{-1em}
\caption{We show the average precision (AP) for object classification on PASCAL VOC 2011 using templates estimated with noise. Even
though the template is created without a dataset, it performs significantly above chance.}
\label{fig:performance}
\end{figure}
\sidecaptionvpos{figure}{b}

The results in Fig.\ref{fig:performance} suggest that biases from the human
visual system do capture some signals useful for classifying objects in real
images. Although the classifiers are estimated using only white noise, in most
cases the templates are significantly outperforming chance, suggesting
that biases from the human visual system may be beneficial
computationally.


\sidecaptionvpos{figure}{b}
\begin{SCfigure}
\centering
\includegraphics[width=0.8\linewidth]{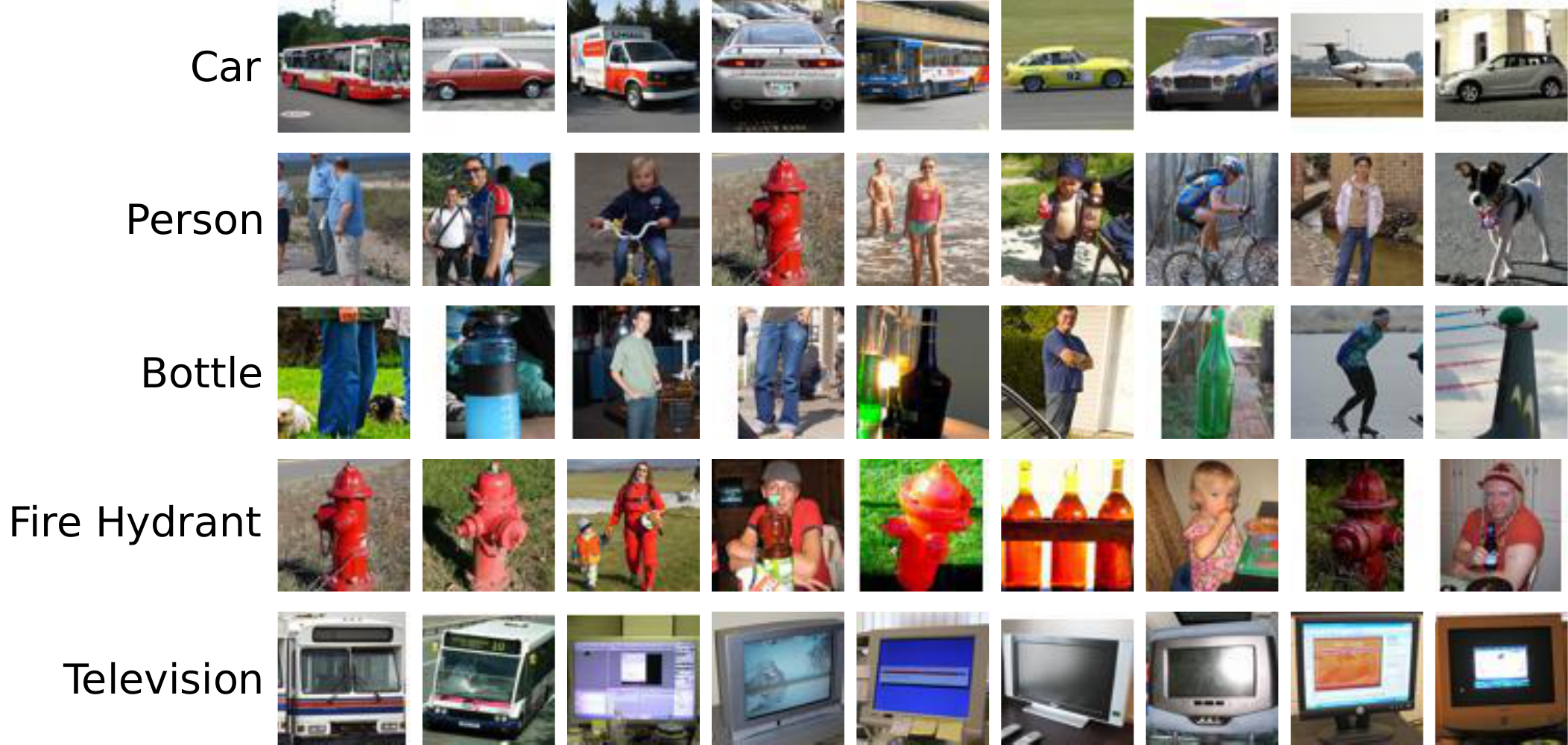}
\caption{We show some of the top classifications from the human biases
estimated with CNN features. Note that real data is not used in building these models.}
\label{fig:topclass}
\end{SCfigure}
\sidecaptionvpos{figure}{c}


\begin{figure}[t]
\centering
\includegraphics[width=0.30\linewidth]{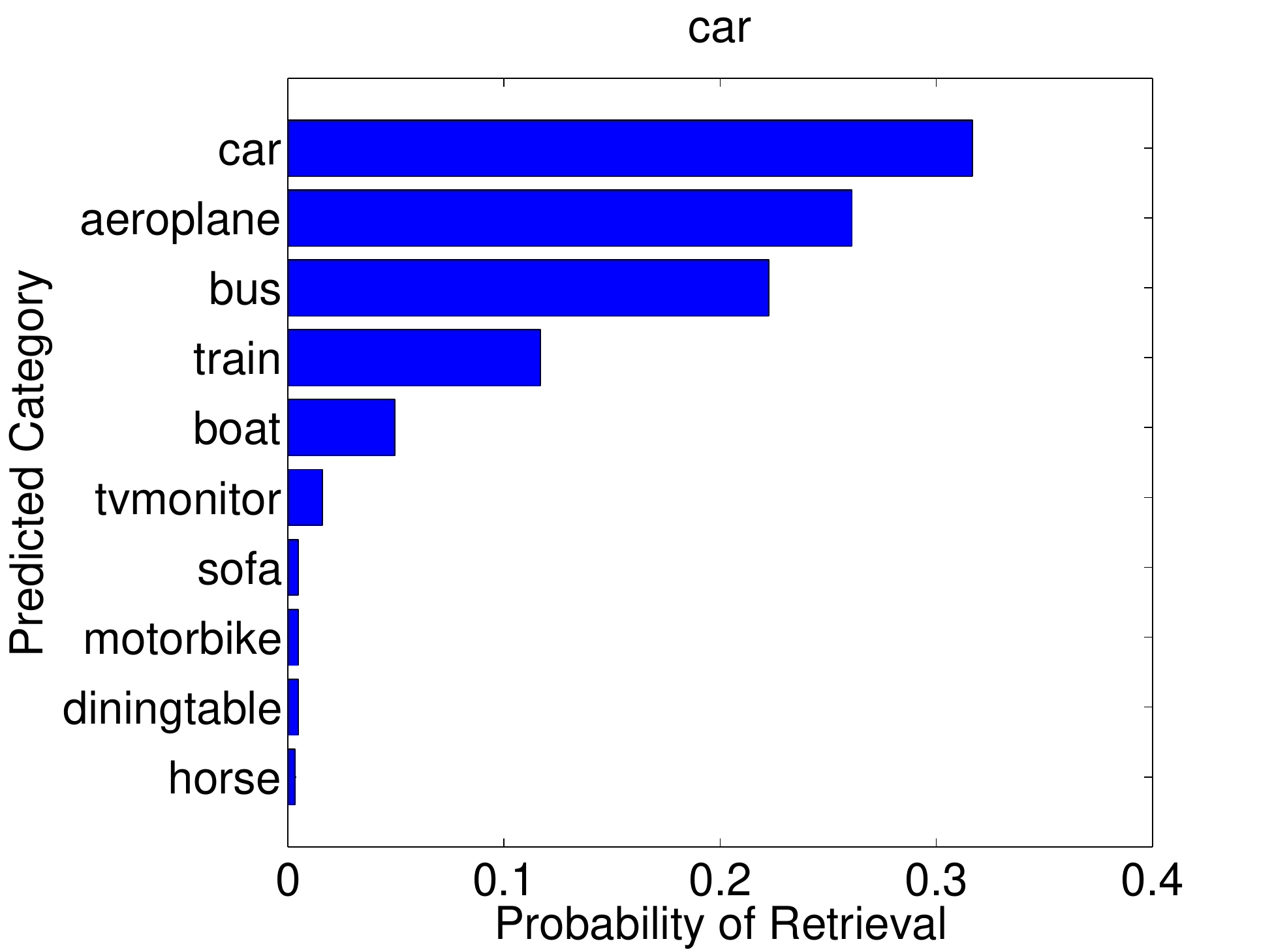}\includegraphics[width=0.30\linewidth]{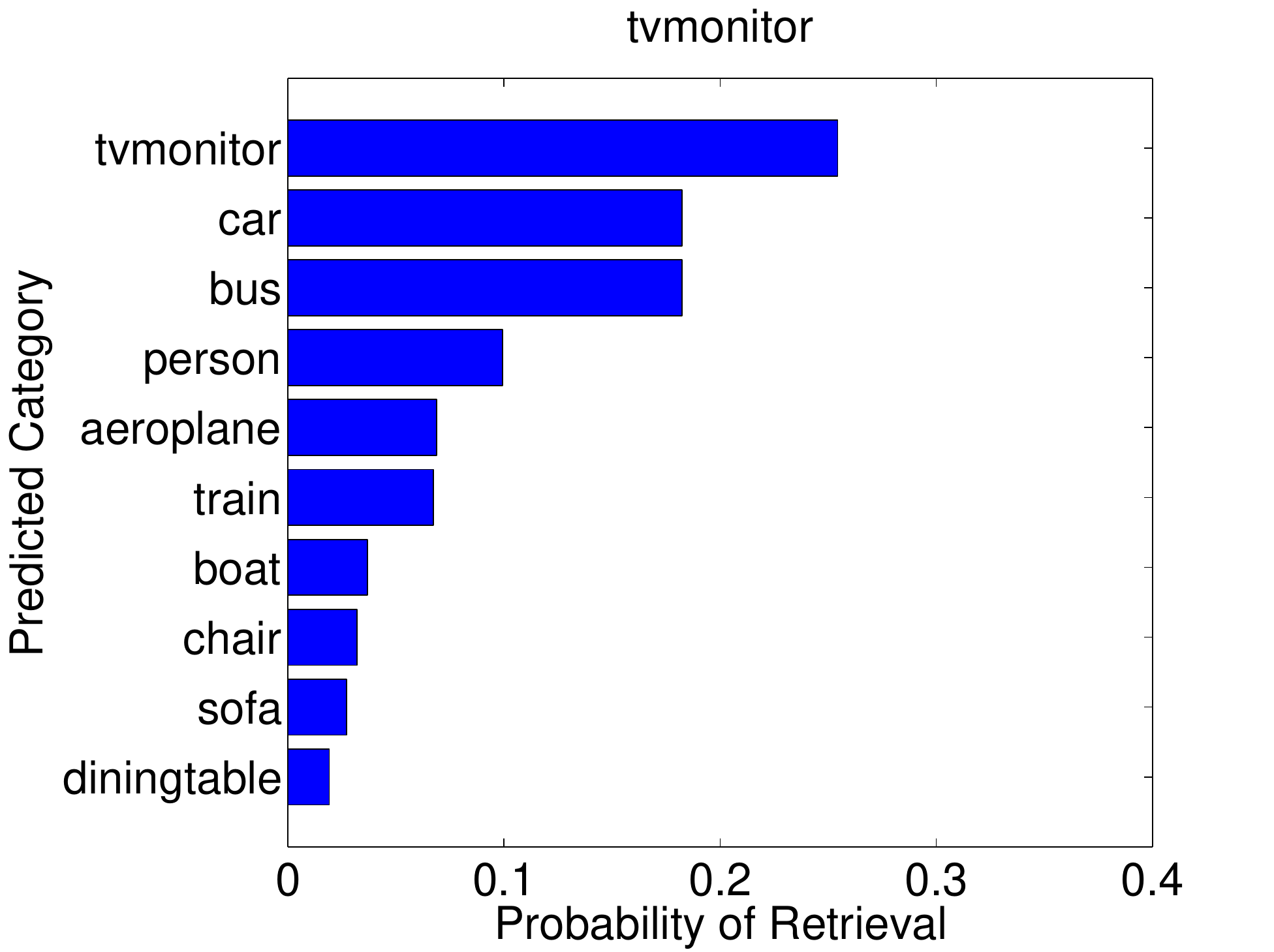}\includegraphics[width=0.30\linewidth]{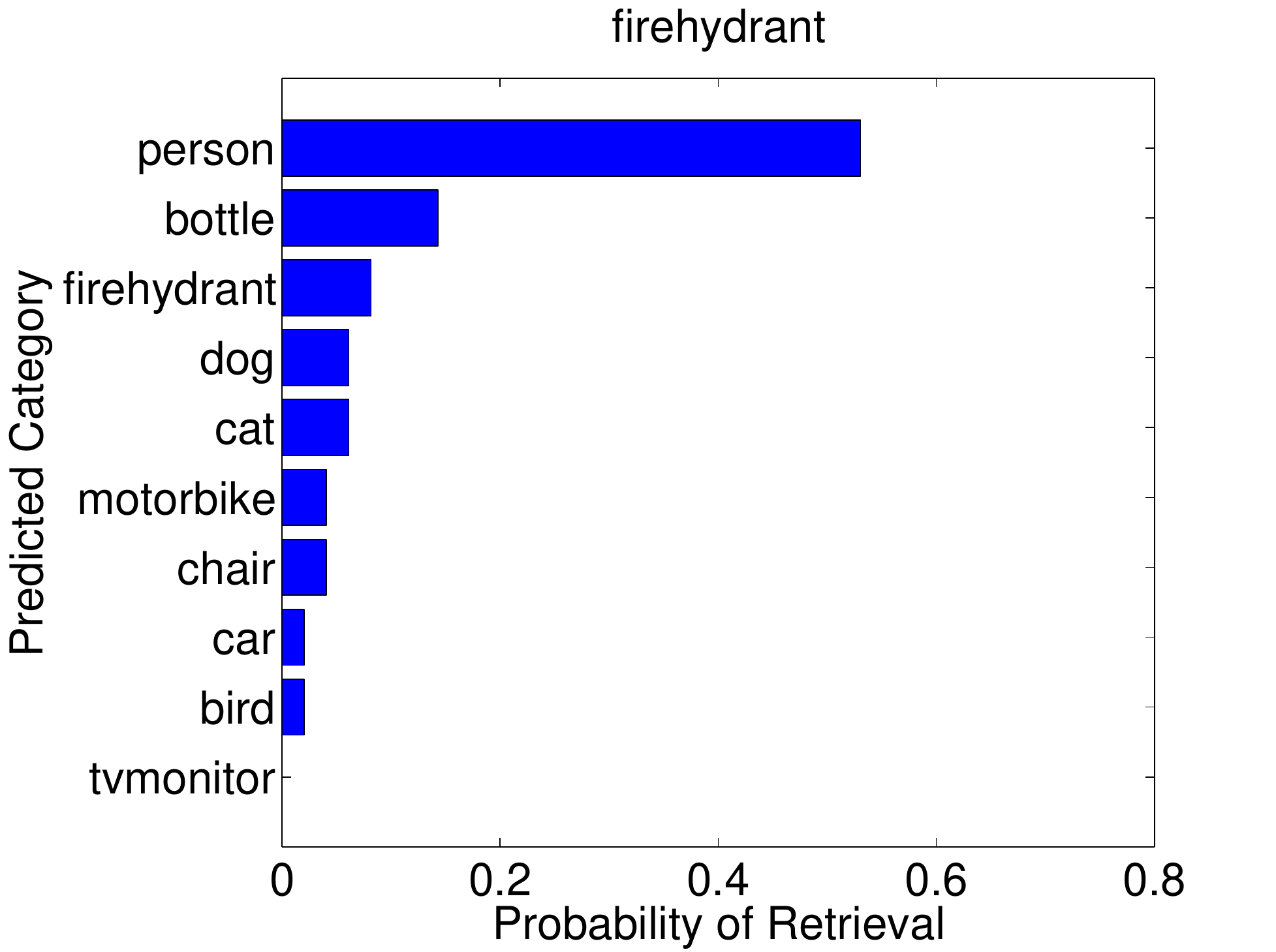}

\caption{We plot the class confusions for some human biases on top
classifications with CNN features. We show only the top 10 classes for
visualization. Notice that many of the confusions may be
sensible, e.g. the classifier for car tends to retrieve vehicles,
and the fire hydrant classifier commonly mistakes people and
bottles.}

\label{fig:confusions}

\end{figure}



Our results suggest that shape is an important bias to
discriminate objects in CNN feature space. Notice how the top classifications in
Fig.\ref{fig:topclass} tend to share the same rough shape by category. For
example, the classifier for person finds people that are upright, and
the television classifier fires on rectangular shapes. The confusions
are quantified Fig.\ref{fig:confusions}: bottles are often
confused as people, and cars are confused as buses. 
Moreover, some templates appear to rely on color as
well. Fig.\ref{fig:topclass} suggests that the classifier for
fire-hydrant correctly favors red objects, which is evidenced by it
frequently firing on people wearing red clothes. The bottle classifier seems to
be incorrectly biased towards blue objects, which contributes to its poor
performance. 

While the motivation of this experiment has been to study whether human biases
are favorable for recognition, our approach has some applications.
Although templates estimated from white noise will likely never be a substitute
for massive labeled datasets, our approach can be helpful for recognizing
objects when no training data is available. Rather, our approach enables us to
build classifiers for categories that a person has only imagined and never
seen. In our experiments, we evaluated on common categories to make evaluation
simpler, but in principle our approach can work for rare categories as well.
We also wish to note that the CNN features used here are trained to classify
images on ImageNet \cite{deng2009imagenet} LSVRC 2012, and hence had access to
data. However, we showed competitive results for HOG as well, which is a
hand-crafted feature, as well as results for a category that the CNN network
did not see during training (fire hydrants).




\iftrue

\section{Learning with Human Biases}
\label{sec:svm}

Our experiments to visualize the templates and use them as object recognition
systems suggest that visual biases from the human visual system provide some
signals that are useful for discriminating objects in real world images. In this section,
we investigate how to incorporate these signals into learning algorithms when there is some training data available. We present an SVM that
constrains the separating hyperplane to have an orientation similar to the human bias
we estimated.


\subsection{SVM with Orientation Constraints}
Let $x_i \in \mathbb{R}^m$
be a training point and $y_i \in \{-1,1\}$ be its label for $1 \le i \le n$. A standard SVM seeks a
separating hyperplane $w \in \mathbb{R}^m$ with a bias $b \in \mathbb{R}$ that
maximizes the margin between positive and negative examples. We wish to add
the constraint that the SVM hyperplane $w$ must be at most $\cos^{-1}(\theta)$ degrees away
from the bias template $c$:
\begin{wrapfigure}[7]{r}{11em}
\centering
\vspace{-0.5em}
\includegraphics[width=10em]{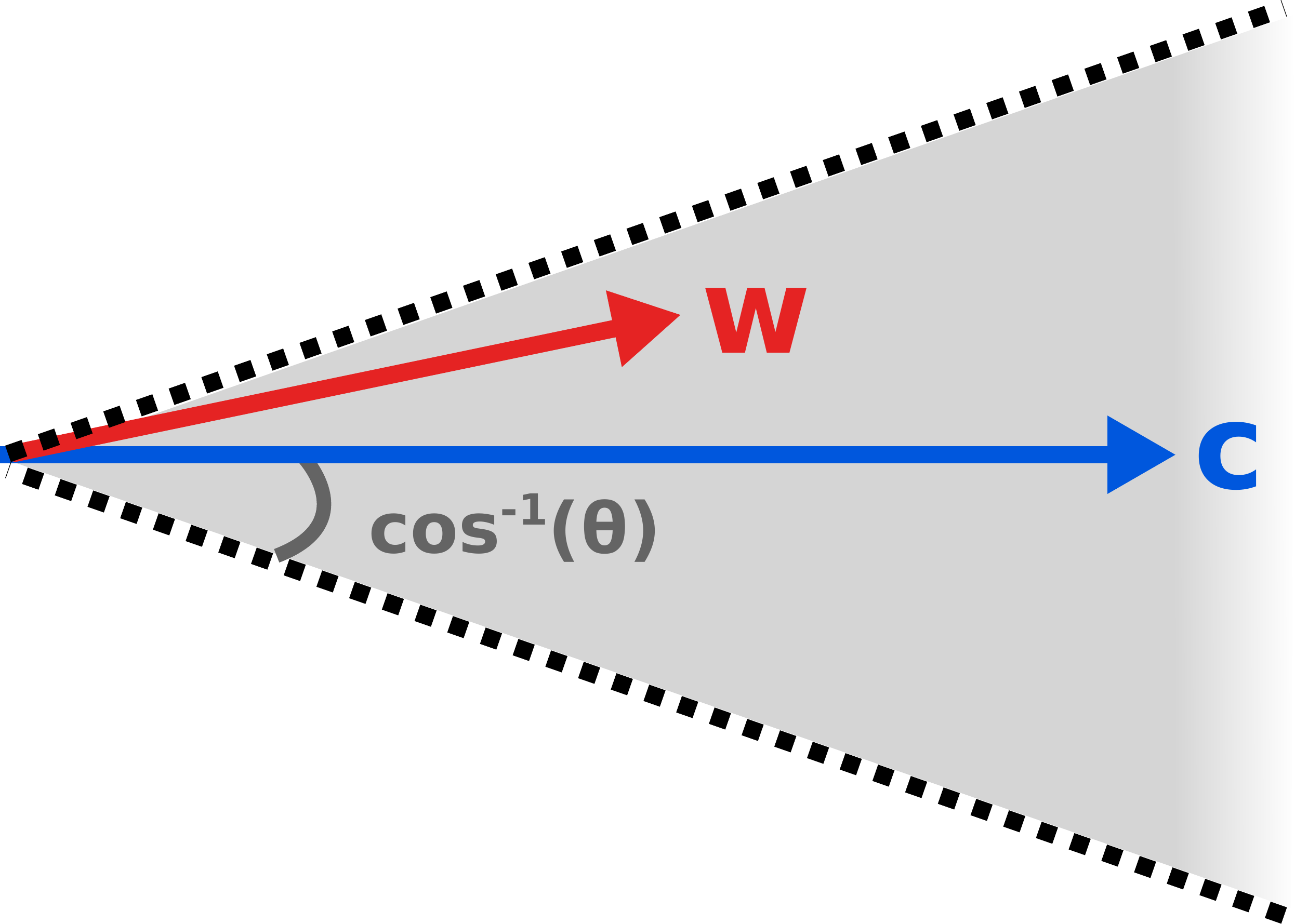}
\caption{}
\label{fig:conic-program}
\end{wrapfigure}
\begin{subequations}
\begin{align}
\min_{w,b,\xi} \frac{\lambda}{2} w^Tw + \sum_{i=1}^n \xi_i \quad \textrm{s.t.} \quad &y_i \left(w^T x_i + b\right) \ge 1 - \xi_i, \; \xi_i \ge 0\\
&\theta \le \frac{w^T c}{\sqrt{w^T w}}
\label{eqn:rsvm-constraint}
\end{align}
\label{eqn:rsvm}%
\end{subequations}
where $\xi_i \in \mathbb{R}$ are the slack variables, $\lambda$ is the regularization hyperparameter, and Eqn.\ref{eqn:rsvm-constraint}
is the orientation prior such that $\theta \in (0,1]$ bounds the maximum angle that
the $w$ is allowed to deviate from $c$. Note that we have assumed, without loss of generality, that $||c||_2 = 1$.
Fig.\ref{fig:conic-program} shows a visualization of this orientation
constraint.  The feasible space for the solution is the grayed hypercone. The
SVM solution $w$ is not allowed to deviate from the prior classifier $c$ by more
than $\cos^{-1}(\theta)$ degrees.

\subsection{Optimization}
We optimize Eqn.\ref{eqn:rsvm} efficiently by writing the objective as a conic
program. 
We rewrite Eqn.\ref{eqn:rsvm-constraint} as $\sqrt{w^T w} \le \frac{w^T c}{\theta}$ and introduce an
auxiliary variable $\alpha \in \mathbb{R}$ such that $\sqrt{w^T w} \le \alpha \le \frac{w^T c}{\theta}$.
Substituting these constraints into Eqn.\ref{eqn:rsvm} and replacing
the SVM regularization term with $\frac{\lambda}{2}\alpha^2$  leads to the conic program:
\begin{subequations}
\begin{align}
\min_{w,b,\xi,\alpha} \frac{\lambda}{2} \alpha^2 + \sum_{i=1}^n \xi_i \quad \textrm{s.t.} \quad &y_i \left(w^T x_i + b\right) \ge 1 - \xi_i, \quad \xi_i \ge 0, \quad \sqrt{w^T w} \le \alpha \\
&\alpha \le \frac{w^T c}{\theta}
\label{eqn:rsvm2-constraint}
\end{align}
\label{eqn:rsvm2}%
\end{subequations}
Since at the minimum $a^2 = w^Tw$, Eqn.\ref{eqn:rsvm2} is equivalent to Eqn.\ref{eqn:rsvm}, but in a standard conic program form.
As conic programs are convex by construction, we can then optimize it efficiently
using off-the-shelf solvers, which we use MOSEK \cite{mosek}. Note that removing Eqn.\ref{eqn:rsvm2-constraint} makes it
equivalent to the standard SVM.
$\cos^{-1}(\theta)$ specifies the angle of the cone. In our
experiments, we found $30^\circ$ to be reasonable. While this angle is not very
restrictive in low dimensions, it becomes much more restrictive as the number
of dimensions increases \cite{li2011concise}.

\subsection{Experiments}




We previously used the bias template as a classifier for recognizing objects
when there is no training data available. However, in some cases, there may be a few
real examples available for learning.
We can incorporate the bias template into learning 
using an SVM with orientation constraints.
Using the same evaluation procedure as the previous section, we compare three
approaches: 1) a single SVM trained with only a few positives and the entire
negative set, 2) the same SVM with orientation priors for $\textrm{cos}(\theta) = 30^\circ$ on the human bias, and 3)
the human bias alone. We then follow the same experimental setup as before.  We
show full results for the SVM with orientation priors in
Fig.\ref{fig:rsvm-results}. In general, biases from the human visual system can
assist the SVM when the amount of positive training data is only a few
examples. In these low data regimes, acquiring classifiers from the human
visual system can improve performance with a margin, sometimes 10\%
AP.

\begin{figure*}[t]
\centering
\begin{tabular}{r | c c | c c | c c}
& \multicolumn{2}{c|}{0 positives} &\multicolumn{2}{c|}{1 positive} & \multicolumn{2}{c}{5 positives} \\
Category & Chance & Human & SVM & SVM+Human & SVM & SVM+Human \\
\hline
car       &  7.3 & \textbf{27.5} & 11.6 & \textbf{29.0} & 37.8 & \textbf{43.5} \\
person    & 32.3 & \textbf{65.6} & 55.2 & \textbf{69.3} & 70.1 & \textbf{73.7} \\
f-hydrant &  0.3 & \textbf{5.9} &  1.7 &  \textbf{7.0 } & \textbf{50.1} & \textbf{50.1}  \\
bottle    &  4.5 & \textbf{6.0} & 11.2 &  \textbf{11.7} & 38.1 & \textbf{38.7} \\
tv &  2.6 & \textbf{23.8} & 38.6 & \textbf{43.1} & 66.7 & \textbf{68.8} \\
\end{tabular}
\caption{We show AP for the SVM with orientation priors for object classification
on PASCAL VOC 2011 for varying amount of positive data with CNN features. All results are means over random subsamples
of the training sets. SVM+Hum refers to SVM with the human bias as an orientation prior.}
\label{fig:rsvm-results}
\end{figure*}

\begin{figure*}[t]
\centering
\subfloat[Train on Caltech 101, Test on PASCAL]{
\includegraphics[width=0.49\linewidth]{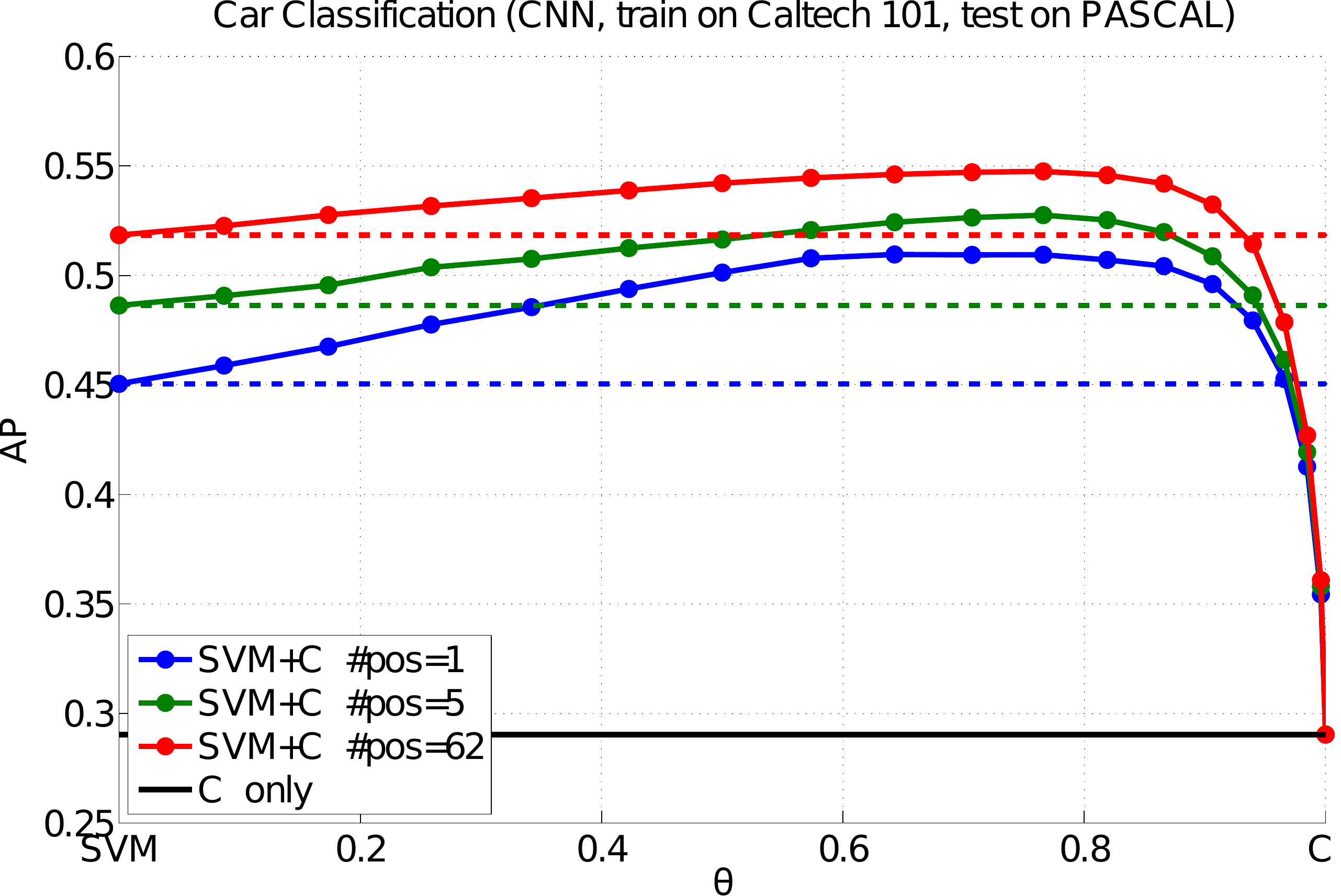}
\label{fig:bias-caltech}
}
\subfloat[Train on PASCAL, Test on Caltech 101]{
\includegraphics[width=0.49\linewidth]{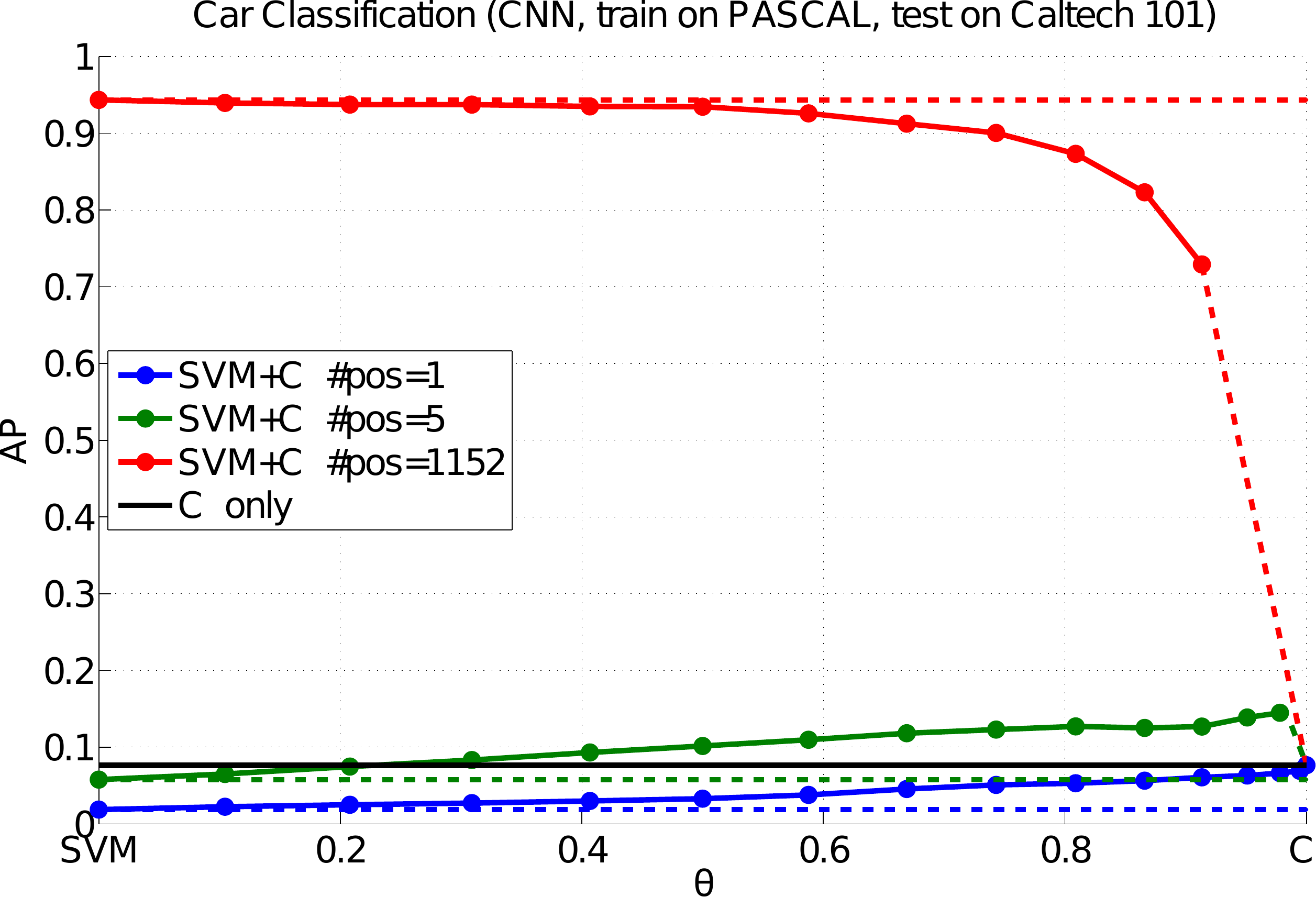}
\label{fig:bias-pascal}
}

\caption{Since bias from humans is estimated with only
noise, it tends to be biased towards the human visual system instead of datasets.  (a) We train an SVM to classify cars on Caltech
101 that is constrained towards the bias template, and evaluate it on
PASCAL VOC 2011. For every training set size, constraining the SVM to the
human bias with $\theta \approx 0.75$ is able to improve
generalization performance. (b) We train a constrained SVM on PASCAL VOC 2011
and test on Caltech 101. For low data regimes, the human bias may help
boost performance.}

\end{figure*}


Furthermore, standard computer vision datasets often suffer from dataset biases that harm cross dataset generalization performance
\cite{torralba2011unbiased,ponce2006dataset}. Since 
the template we estimate is biased by the human visual system and not datasets (there is no dataset),
we believe our approach may help cross dataset generalization.
We trained an SVM classifier with CNN features to recognize cars on Caltech 101
\cite{fei2006one}, but we tested it on object classification with PASCAL VOC
2011.  Fig.\ref{fig:bias-caltech} suggest that, by constraining the SVM to be
close to the human bias for car, we are able to improve the generalization
performance of our classifiers, sometimes over 5\% AP. We then tried the
reverse experiment in Fig.\ref{fig:bias-pascal}: we trained on PASCAL VOC 2011,
but tested on Caltech 101. While PASCAL VOC provides a much better sample of
the visual world, the orientation priors still help generalization performance
when there is little training data available.  These results suggest
that incorporating the biases from the human visual system may
help alleviate some dataset bias issues in computer vision.

\fi

\section{Conclusion}

Since the human visual system is one of the best recognition systems, we
hypothesize that its biases may be useful for visual understanding. In this
paper, we presented a novel method to estimate some biases that people have for
the appearance of objects. By estimating these biases in state-of-the-art
computer vision feature spaces, we can transfer these templates into a machine,
and leverage them computationally. Our experiments suggest biases from the
human visual system may provide useful signals for computer vision systems,
especially when little, if any, training data is available.

%

{\small
\paragraph{Acknowledgements:}
We thank Aditya Khosla for important discussions, and Andrew Owens and
Zoya Bylinskii for helpful comments.
Funding for this research was partially supported by a Google PhD Fellowship to CV,
and a Google research award and ONR MURI N000141010933 
to AT. 
}

{
\small
\bibliographystyle{ieee}
\bibliography{main}
}

\end{document}